\documentclass[letterpaper, 10 pt, journal, twoside]{IEEEtran} 
\usepackage[top=0.792in, bottom=0.75in, right=0.687in, left=0.687in]{geometry}
\usepackage{microtype}

\usepackage{graphicx}
\usepackage{booktabs} 
\usepackage{hyperref}
\usepackage{url}            
\usepackage{amsfonts}       
\usepackage{nicefrac}       
\usepackage{microtype}      
\usepackage[utf8]{inputenc}
\usepackage{amsmath}
\usepackage{gensymb}
\usepackage{graphicx}
\usepackage{wrapfig}
\usepackage{float}
\usepackage[font=footnotesize,justification=justified]{caption}
\usepackage{mathtools}
\usepackage{verbatim}
\usepackage{tikz}
\usepackage{enumitem}  
\usepackage{cancel}
\usepackage{times} 
\usepackage{amssymb,longtable,calc}
\usepackage{graphbox}
\usepackage{mathptmx}
\usepackage{xspace}
\usepackage{latexsym}
\usepackage{multicol}
\usepackage{multirow}
\usepackage{color,comment,rotating,url,xspace} 
\usepackage{subcaption}
\usepackage{tabularx}
\usepackage{cite}
\usepackage[ruled,vlined]{algorithm2e}  
\newcolumntype{Y}{>{\centering\arraybackslash}X}
\newcommand{\figref}[1]{Fig.~\ref{#1}}

\newcommand{\secref}[1]{Sec.~\ref{#1}}

\makeatletter
\let\@minipagerestore=\raggedright
\makeatother
\newlength{\mylen}
\setbox1=\hbox{$\bullet$}\setbox2=\hbox{\tiny$\bullet$}
\setlist[itemize]{itemsep=0mm, topsep=2pt}

\title{Learning Task-Oriented Grasping from Human Activity Datasets}

\begin{document}

\author{Mia~Kokic$^{1}$, Danica~Kragic$^{1}$,~\IEEEmembership{Fellow,~IEEE} and Jeannette~Bohg$^{2}$,~\IEEEmembership{Member,~IEEE}%
\thanks{Manuscript received: September 10, 2019; Revised: January 4, 2020; Accepted: January 29, 2020.}
\thanks{This paper was recommended for publication by Editor Hong Liu upon evaluation of the Associate Editor and Reviewers' comments. This work was partially supported by the Wallenberg AI, Autonomous Systems and Software Program (WASP) funded by the Knut and Alice Wallenberg Foundation.
Toyota Research Institute ("TRI") provided funds to assist the authors with their research but this article solely reflects the opinions and conclusions of its authors and not TRI or any other Toyota entity.}%
\thanks{$^{1}$Mia Kokic and Danica Kragic are with the RPL, EECS, KTH, Stockholm, Sweden {\tt \{mkokic|dani\}@kth.se}}%
\thanks{$^{2}$Jeannette Bohg is with the Computer Science Department, Stanford University, CA, USA. {\tt bohg@stanford.edu}}%
\thanks{Digital Object Identifier (DOI): see top of this page.}
}

\markboth{IEEE Robotics and Automation Letters. Preprint Version. Accepted February, 2020}
{Kokic \MakeLowercase{\textit{et al.}}: Learning Task-Oriented Grasping from Human Activity Datasets} 

\maketitle

\begin{abstract}
We propose to leverage a real-world, human activity RGB dataset to teach a robot {\em Task-Oriented Grasping} (TOG).
We develop a model that takes as input an RGB image and outputs a hand pose and configuration as well as an object pose and a shape. 
We follow the insight that jointly estimating hand and object poses increases accuracy compared to estimating these quantities independently of each other. 
Given the trained model, we process an RGB dataset to automatically obtain the data to train a TOG model. This model takes as input an object point cloud and outputs a suitable region for task-specific grasping. 
Our ablation study shows that training an object pose predictor with the hand pose information (and vice versa) is better than training without this information. Furthermore, our results on a real-world dataset show the applicability and competitiveness of our method over state-of-the-art. Experiments with a robot demonstrate that our method can allow a robot to preform TOG on novel objects.
\end{abstract}


\vspace{-5px}
\section{Introduction}
\label{sec:intro}

\IEEEPARstart{K}{nowing} accurate poses and shapes of hands and objects during everyday manipulation tasks can provide cues for teaching robots to grasp objects in a task-oriented manner. For example, humans typically grasp knives at their handle to cut with the blade. 
\begin{figure}[h!]
\centering
\includegraphics[width=0.35\textwidth]{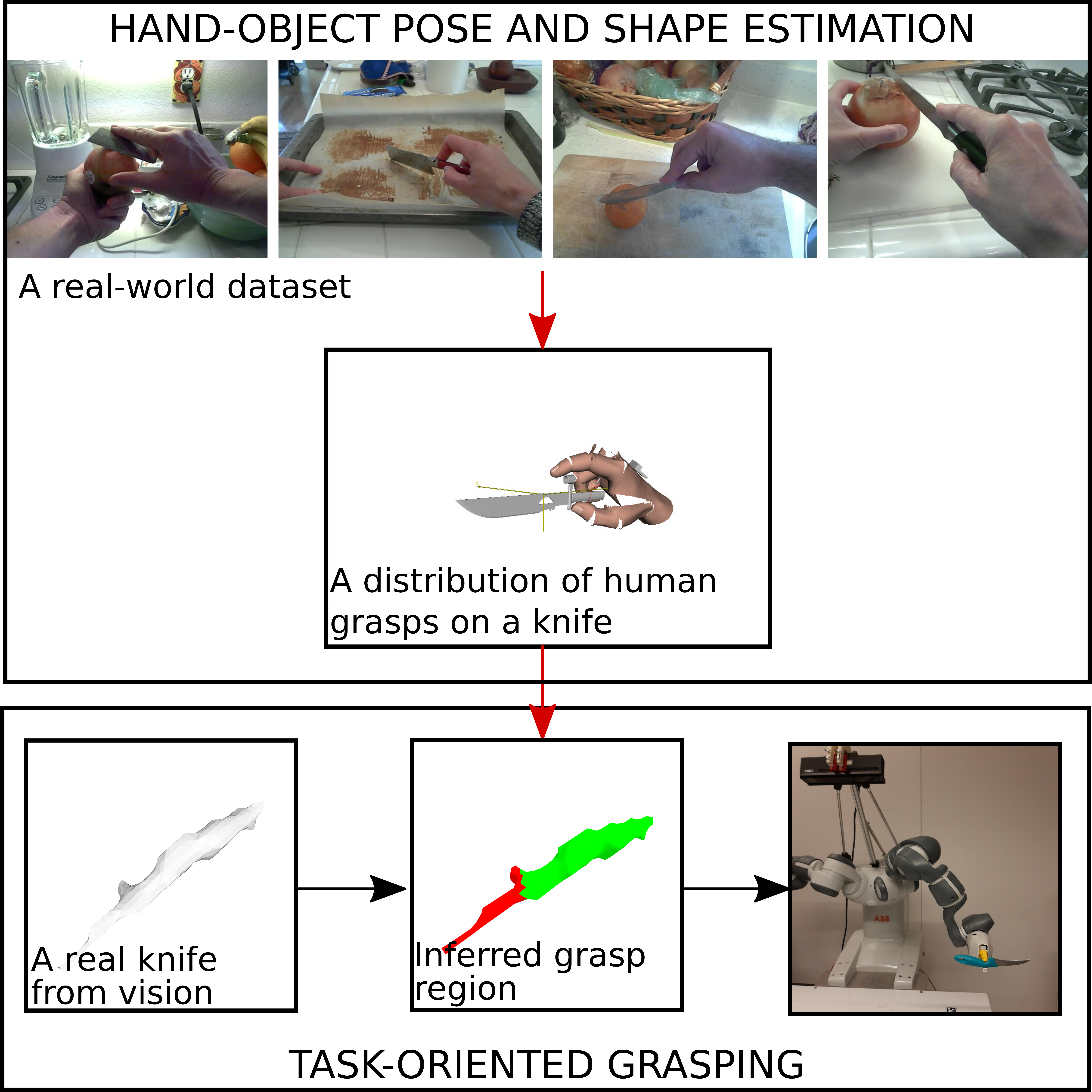}
\caption{Top to bottom: RGB images of human grasps on {\em knives} for cutting; the distribution of human grasps on a knife as a results of processing RGB images; a knife recorded with an RGB-D camera, suitable grasping area, a task-oriented grasp with a robot. Red arrows indicate our contributions: 1) estimating hand and object poses and shapes from RGB images and 2) using that data to teach a robot TOG.}
\label{fig:intro}
\end{figure}
There are two main approaches for learning {\em Task-Oriented Grasping\/} (TOG). They involve either manually labeling contact regions on objects for a certain task~\cite{kokic2017affordance, antonova2018global, detry2017taskoriented} or executing task motion trajectories and scoring the grasps based on the action outcomes~\cite{fang2018learning}. 
A major bottleneck with both approaches is the tedious data collection that requires either manual labeling of contact regions or designing task trajectories. 

A way to circumvent this is to leverage rich human activity datasets that contain images and videos of humans manipulating objects~\cite{garcia2018first, li2015delving, li2018eye}. The caveat is that these datasets are often 2D and lack annotations that could facilitate the inference of a 6D grasp pose which is central to grasping. A potential solution is to lift this data to 3D which is a challenging problem in itself. While there has been a lot of work on pose and shape estimation of either hands~\cite{zimmermann2017learning, mueller2017real, mueller2018ganerated, ge20193d} or objects~\cite{xiang2017posecnn, sundermeyer2018implicit, balntas2017pose} from RGB images, research on jointly estimating hands and objects is scarce~\cite{hasson2019learning, tekin2019h+, kokic2019learning, romero2013non}.
The challenges are: 1) strong occlusions from both hands and objects and 2) tedious annotations of real images with hand and object information. 
While the first problem has been addressed in the past~\cite{kokic2019learning, hasson2019learning, tekin2019h+} by devising algorithms that estimate hands and objects in conjunction, the lack of large real-world, annotated datasets impedes the generalization capacity of these algorithms to novel objects.
To address this,~\cite{hasson2019learning} generated a large synthetic dataset of a hand grasping objects on which the authors train a model to reconstruct hand-object pairs in 3D. However, the results often lack sufficient detail especially for the objects. Furthermore, synthetic datasets suffer from the sim-to-real gap.

In this paper, we present an approach for learning TOG by processing a real-world dataset of RGB images showing humans manipulating various objects. This provides us with distributions of task-oriented grasps on objects which can be used to teach a robot how to grasp novel objects in a task-oriented manner (see~\figref{fig:intro}). 
For processing the dataset, we develop a method that predicts a hand pose and a configuration as well as an object pose and a shape from a single RGB image. 
We exploit the notion that jointly estimating hands and objects can be beneficial and devise a network architecture that internally couples them. In the experiment section, we quantitatively evaluate the proposed model by reporting the effect of using a hand to estimate an object pose and vice versa. 
Our method shows competitive results in comparison to state-of-the art~\cite{hasson2019learning} while being more detailed in object shape estimation.
To transfer this knowledge to a robot, we train a CNN that predicts if a grasp is suitable for a task on a novel object.
We demonstrate the applicability of our approach by executing task-oriented grasps with a real robot on novel objects from a known category. 
In summary, we make the following contributions:
\begin{enumerate}[wide, labelwidth=!, labelindent=0pt]
    \item A framework for estimating hand pose and configuration as well as object pose and shape from a single RGB image. 
    \item Competitive results to state-of-the-art~\cite{hasson2019learning} and generalization to novel object instances from three categories. 
    \item TOG models learned from a challenging real-world dataset of humans manipulating objects. 
    \item Real robot demonstrations of task-oriented grasps on previously unseen object instances from a given category. 
\end{enumerate}
\vspace{-5px}
\section{Related Work}
\label{sec:relatedwork}
\subsection{Modeling Hand-Object Interactions}
Tekin et al.~\cite{tekin2019h+} propose a CNN that jointly predicts hand and object poses. Their model was trained on a real-world {\em First-Person Hand Action Benchmark (FHB)} dataset~\cite{garcia2017firstperson} which contains hand annotations per image. However, only 4 objects are annotated with poses. To address the lack of object annotations,~\cite{hasson2019learning} generate an RGB dataset of synthetic hands and objects on different backgrounds and train a CNN that reconstructs meshes of hand-object pairs from an image. 
However, reconstruction often requires a strong regularization of the shape due to the large dimensionality of the problem. Thus, the output objects in~\cite{hasson2019learning} often have blob-like shapes which makes it difficult to discern shape details that may be crucial for learning TOG. 
For example, when learning to grasp a knife for cutting, it is important to know where the blade is. 
While \cite{tekin2019h+} and~\cite{hasson2019learning} yield promising results for the hand estimation,~\cite{tekin2019h+} lacks generalization power over object shapes due to only four objects being annotated and~\cite{hasson2019learning} lacks sufficient detail in reconstructed objects and hands.

In this paper, we build upon our previous work~\cite{kokic2019learning} where we train a CNN that predicts an object pose and a shape from an RGB image. It uses information about the hand as additional input to the CNN which we showed to significantly improve the results of object pose estimation. To obtain the information on the hand, we run an existing hand pose detector~\cite{zimmermann2017learning} on the input RGB image. However, if the hand pose estimate is inaccurate, this can negatively influence the object pose estimate. This is often the case, since the dataset used for training of~\cite{zimmermann2017learning} does not contain a lot of samples with object occlusions. 
To address this limitation, in this work, we train a CNN that jointly predicts the hand and the object from an RGB image. It is trained on a combination of the real-world {\em FHB\/} dataset and our own synthetic dataset. {\em FHB\/} contains dense hand annotations and the hand is partially occluded by the objects. This ensures generalization over different hand poses.
The synthetic dataset contains dense object pose and shape annotations for a wide variety of objects that are partially occluded by the hand. This ensures generalization to novel objects and poses. 
As in our previous work, our method outputs an object pose and a shape descriptor which is used to retrieve the most similar mesh from a set of meshes for that category. We extend our previous work by also predicting the hand pose and joint angles.
We show in the experiments that, in contrast to~\cite{hasson2019learning}, our method yields more precise estimates of hands and objects.

\subsection{Task-Oriented Grasping}
The problem of grasping an object in a manner that allows for an execution of a task is known as {\em Task-Oriented Grasping}~\cite{song2015task, kokic2017affordance, fang2018learning, bohg2013data} and is more challenging than simply grasping an object to hold it stably. 
The first challenge is to find grasps that have the right trade-off between stable and task-compliant. 
The second challenge pertains to data collection that would enable all-purpose algorithms that can deal with variety of objects and tasks. 
To collect the data, authors often manually label regions on objects that are suitable for TOG. 

For example,~\cite{kokic2017affordance} train a CNN that predicts part affordances, which are used to formulate constraints on the location and orientation of the gripper. Constraints are given to an optimization-based grasp planner, which executes the grasps. 
Similarly,~\cite{detry2017taskoriented} train task-oriented CNNs to identify regions a robot is allowed to contact to fulfill a task. 
Another approach is to learn TOG from self-supervision. Fang et al.~\cite{fang2018learning} generate task-specific motion trajectories with objects in simulation and used them to score grasps on objects for two tasks. Scoring is based on whether the task succeeded or not. 
The challenge with these approaches is to either manually label contact regions or generate the task trajectories which is often expensive. We propose a model for processing a real-world dataset of RGB images that, when lifted to 3D allows us to obtain annotations automatically.

\begin{figure*}[h!]
\centering
\includegraphics[width=0.7\textwidth]{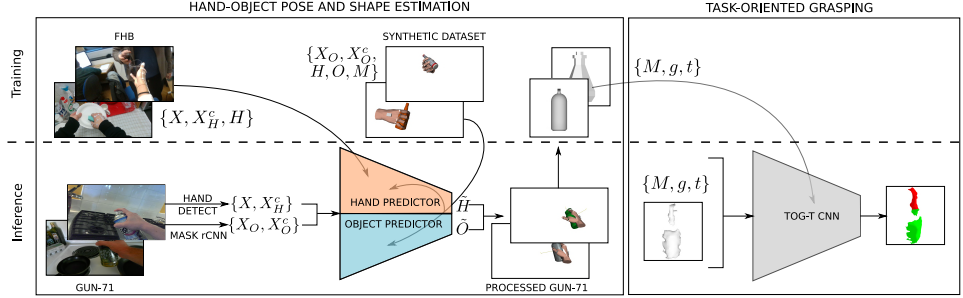}
\caption{\label{fig:pipe} On the left we show a model for the hand-object pose and shape estimation and on the right, the model for TOG. Above the dashed line are the datasets used for training of these models (curved line connects a network with the dataset it was trained on). Below the dashed line is the information flow during inference: A real-world dataset GUN-71 is pre-processed and fed to the hand-object pose and shape estimation network. This yields meshes annotated with grasps for a task which are used to train the {\em TOG-T} network. When presented with a real object and a task, the {\em TOG-T} network predicts task-suitable regions for grasping.}
\end{figure*}
\vspace{-5px}
\section{From RGB Images to Task-Oriented Grasps}
\label{sec:method}
We develop a method that enables a robot to execute task-oriented grasps on novel objects from a known category. To this end, we learn a category specific model from examples of task-oriented grasps that are mined from an RGB dataset of human manipulation actions. We divide the problem into two subproblems. The first subproblem pertains to processing an RGB dataset to obtain 3D hand-object pairs. The second subproblem pertains to executing task-oriented grasps by using the data generated in the previous stage.

To address the first subproblem (\figref{fig:pipe}, left), we train a CNN that takes as input an RGB image of a hand and an object and outputs the hand pose and configuration (we use terms hand configuration and hand shape interchangeably) and the object pose and shape descriptor used to retrieve the most similar looking mesh to the object in the image. 
To address the second subproblem (\figref{fig:pipe}, right), executing task-oriented grasps, we run the network on the RGB datasets of human manipulation actions and obtain statistics on the relative hand poses with respect to retrieved object meshes. This data is then used to transfer the task-oriented grasp experience to a robot which is presented with a novel object of a given category. We transfer this knowledge by training a CNN that takes as input the novel object in a form of a raw point cloud obtained from vision, a task and a grasp and predicts whether the grasp is suitable for the task or not. 
\vspace{-3px}
\subsection{Hand-Object Pose and Shape Estimation}
\label{sec:hops}
Given an RGB image $X$ of a human hand holding an object, our goal is to predict a 6D pose and a configuration of the hand $H = (\theta_{H}, \beta)$ as well as a pose and a shape of the object $O = (\theta_{O}, M)$. 
A pose of a hand or an object is parameterized with a transformation matrix $\theta$ which describes a position $p \in \mathbb{R}^3$ and an orientation $W \in \mathbb{R}^{3\times3}$ of the hand or the object in the camera frame (we use subscripts $H$ and $O$ to indicate either hand or object pose).
The configuration of the hand contains $21$ joint angles $\beta \in \mathbb{R}^{21}$ and the shape of the object $M$ is a polygonal mesh. 

\subsubsection{Hand}
\label{sec:hops_hand}
We estimate the hand pose and configuration in two steps. The first step where we obtain the initial hand predictions does not include an information about an object and the second step uses object pose to refine the initial hand predictions. 
We get the initial hand estimates by learning two functions. The first one takes as input an RGB image $X$ and predicts the hand position $\tilde{p}_H$ in the camera frame. The second function takes as input a crop, i.e., a minimum bounding box around the hand $X_{H}^{c}$ and predicts the rotation matrix $\tilde{W}_H$ of the hand's coordinate frame with respect to the camera as well as the hand configuration $\tilde{\beta}$. Formally, we learn two mappings:
\begin{align}
    h_p: &X \mapsto \tilde{p}_H \label{eq:trans_h} \\
    h_{W, \beta}: &X_{H}^{c} \mapsto(\tilde{W}_H, \tilde{\beta}) \label{eq:rot_h}
\end{align}
We use these quantities to estimate the object pose. Once we obtain the object pose estimates, position $\tilde{p}_O$ and orientation $\tilde{W}_O$, we can in turn use them to improve on initial hand estimates by predicting the hand values with the object information. 
Formally:
\begin{align}
    {h_p}': &(X, \tilde{p}_O) \mapsto \tilde{p_H}' \label{eq:trans_h_} \\
    {h_{W, \beta}}': &(X_{H}^{c}, \tilde{W}_O) \mapsto(\tilde{W_H}', \tilde{\beta}) \label{eq:rot_h_}
\end{align}

\subsubsection{Object} 
\label{sec:hops_obj}
We estimate an object position and orientation by learning two functions. The first takes as input an RGB image showing an object segmented from the background $X_{O}$ (obtain with \emph{Mask R-CNN}~\cite{he2017mask}) and a hand position $\tilde{p}_H$ from~\eqref{eq:trans_h} and outputs an object position estimate $\tilde{p}_O$. The second function takes as input an object crop $X_{O}^{c}$ and a hand rotation matrix $\tilde{W}_H$ from~\eqref{eq:rot_h} and outputs an estimated object rotation matrix $\tilde{W}_H$. This yields two functions:
\begin{align}
    o_p: &(X_O, \tilde{p}_H) \mapsto \tilde{p}_O \label{eq:trans_o} \\
    o_W: &(X_{O}^{c}, \tilde{W}_H) \mapsto \tilde{W}_O \label{eq:rot_o}
\end{align}
The object pose estimation problem benefits from knowing the hand pose in several ways. First, the hand provides cues about object position even when the object is occluded by the hand. Second, the hand orientation provide cues about object orientation, e.g., a knife is often grasped at the handle such that the blade is pointing downwards.

We model the object shape estimation as a retrieval problem. Namely, we aim at finding the most similar mesh (among all meshes in a category) to the object in the image. We do this by generating a joint embedding space for shape and image data points (same as in our previous work~\cite{kokic2019learning}). It ensures that an object shape and image features are mapped close together in the embedding space if they belong to the same object and distant otherwise. During inference, this allows us to retrieve the most similar mesh to the object in the image by querying the nearest neighbor in the embedding space.

Formally, the third function takes as input the crop $X_O^c$ and outputs a shape feature vector $\delta(X_{O}^{c})$ which is used to retrieve the most similar looking mesh during inference. 
\begin{align}
    o_M: &X_{O}^{c} \mapsto \tilde{\delta(X}_{O}^{c}) \label{eq:shape_o}
\end{align}

\vspace{-3px}
\subsection{Task-Oriented Grasping}
\label{sec:tog}
As stated in~\secref{sec:method} we want to mine large datasets of RGB images showing hands interacting with objects to learn TOG. To demonstrate this, we run our network for hand-object pose and shape estimation on a realistic RGB dataset depicting hands in interaction with objects.
For each object category, we process all the images in the dataset which yields a set of mesh models annotated with human grasps for a certain task. 
\begin{figure}
\centering
\includegraphics[width=0.25\textwidth]{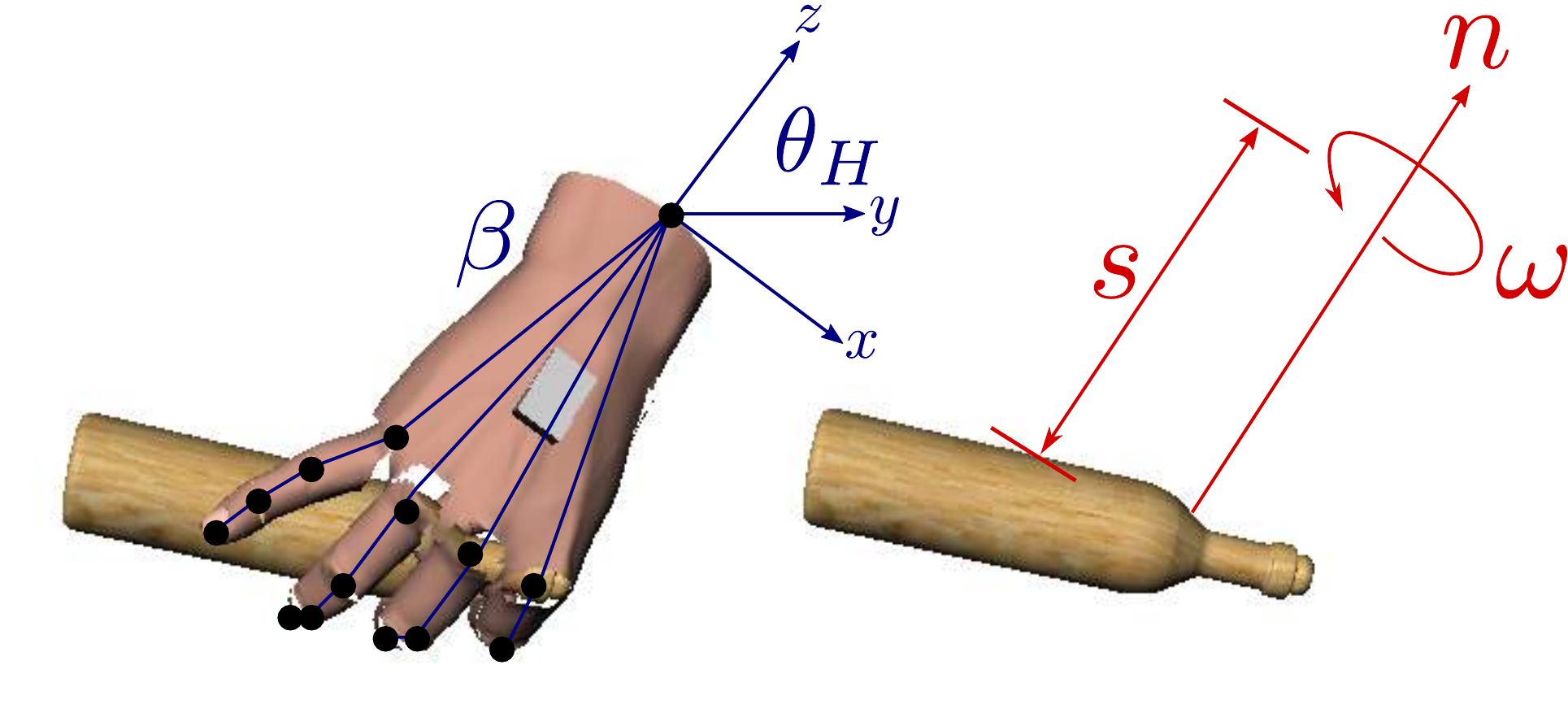}
\caption{Human and robot grasp representations. Black: 3D joint positions. Blue: hand pose $\theta_H$ and configuration $\beta$. Red: Grasp representation used for TOG. It consist of: 1) an approaching normal $n \in SE(3)$, 2) a wrist rotation around the approaching normal, i.e., a roll $\omega \in \mathbb{R}$ and 3) an offset from the object surface $s \in \mathbb{R}$.
A normal $n$ is parallel to the object surface normal and aligned with the $z$-axis of the human hand. For definition of SE(3) see~\cite{cederberg2013course}}
\label{fig:grasp_param} 
\end{figure}
To ensure generalization to previously unseen objects, we train a CNN which we call {\em TOG-T}, that takes as input an object in the form of a binary voxel grid $M_V$, a grasp $g = (n, \omega, s)$ (which we construct from a human hand pose, see~\figref{fig:grasp_param}; we use this representation of a grasp to ensure the transfer to a parallel gripper), and a task $t$. It outputs a probability that the grasp is suitable for the task. We then threshold these values to segment the object into suitable/unsuitable regions for approaching and grasping. The area that is suitable for grasping reflects the human grasp distribution, however, only a subset of grasps in the area are also stable for a specific gripper. 
To address the stability issue, we leverage our previous work where we trained a CNN to predict a stability score of a two-finger antipodal grasp, which we call {\em TOG-S} network~\cite{antonova2018global}. Similarly to {\em TOG-T} network, it takes as input $M_V$ and $g$ and outputs a stability score between $0$ and $1$. We consider those grasps valid that have a stability score $\geq 0.5$ and are deemed suitable for the task.
\vspace{-5px}
\section{Datasets}
\label{sec:datasets}
In this paper, we consider three categories of objects: \emph{bottles, knives} and \emph{spoons}. We use two datasets for training our models: the \emph{synthetic dataset} is used for object pose and shape; the \emph{First-person hand benchmark} (FHB)~\cite{garcia2018first} is used for hand pose and configuration. For testing the model and collecting the task-oriented grasps, we use GUN-71~\cite{rogez2015understanding}, an egocentric dataset of a variety of real hand-object pairs. 

The \textbf{synthetic dataset} consists of approximately $100$ meshes per category taken from ShapeNet and ModelNet40. To generate images and their labels, we use GraspIt!~\cite{miller2004graspit} which has a realistic human hand model. Specifically, we import an object mesh $M$ in a random pose. Then, we sample a grasp from a predefined set $h_o \in H_O$ which we generated off-line per object with the EigenGrasp planner~\cite{EigenGrasp}. Finally, we apply the grasp with the human hand model. We store RGB images $X_O$ and $X_{O}^{c}$, both obtained by masking the hand with white pixels. These images are labeled with the hand pose $\theta_H$ and joint configuration $\beta$.

During the generation of a grasping set $H_O$, we reject the grasps that do not reflect a realistic distribution of human grasps for that category, e.g., we discard the grasps on a blade of a knife. Note, that the remaining grasps are not necessarily task-oriented. They simply reflect the most common object part that humans grasp. Making the grasps task-oriented would require manual inspection which we circumvent by automatically processing the rich RGB dataset.

\textbf{FHB} is an egocentric dataset that contains real RGB-D videos of humans manipulating objects. We decided to train the hand predictor on this data instead of the synthetic dataset by~\cite{hasson2019learning} or our previous work~\cite{kokic2019learning} to avoid the sim-to-real gap.
Each frame is annotated with 3D joint positions relative to the camera frame. Ground truth meshes and pose labels are provided for only 4 of 26 objects. 
We pre-process the data for training the hand predictor by first scaling the joint positions to comply with the human hand model in GraspIt!. Then, via inverse kinematics, we obtain $\theta_H$ and $\beta$ for each RGB frame. Video sequences yield 105,459 images $X$. We also crop these to obtain $X_{H}^{c}$. 

\textbf{GUN-71} is a real-world egocentric dataset which we process to collect the data for TOG. It consists of RGB images of hands manipulating a plethora of different objects from many categories. No annotations, except from the grasp type shown in the image are available.

\vspace{-5px}
\section{Training and Inference}
\label{sec:trinf}

\vspace{-3px}
\subsection{Hand-Object Pose and Shape Estimation}
\label{sec:inf-hops}
\subsubsection{Network Architecture} 
\begin{figure}
\centering
\begin{minipage}{.47\textwidth}
\includegraphics[width=\textwidth]{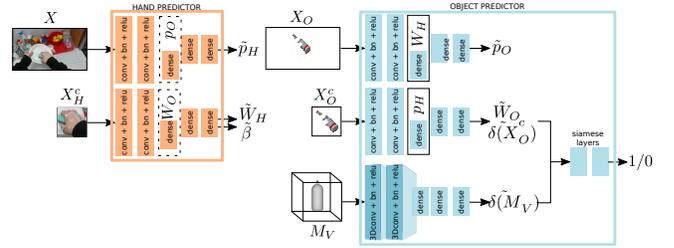}
\caption{Hand-object pose and shape estimation network. Hand predictor takes as input $X$ and $X_{H}^{c}$ and outputs $p_H$, $W_H$ and $\beta$. Object predictor takes as input $X_{O}$, $X_{O}^{c}$ and $M_V$ and outputs $p_O$, $W_O$, $\delta(X_{O}^{c})$ and $\delta(M_V)$. Dotted lines in the hand predictor indicate training with an without an object. Color encoding corresponds to hand (orange) and object predictor (blue) in~\figref{fig:pipe}.}
\label{fig:cnn} 
\end{minipage}
\end{figure}
The network architecture is split into a hand predictor and an object predictor (see~\figref{fig:cnn}). The hand predictor (orange), predicts the hand pose $\theta_H$ and joint configuration $\beta$. The object predictor (blue) predicts the object pose $\theta_O$ and features of the image crop $\delta(X_{O}^{c})$. 

To train the object pose estimator (Eq.~\ref{eq:trans_o} and \ref{eq:rot_o}), we use object and hand annotations from our synthetic dataset. To train the hand pose estimator (Eq.~\ref{eq:trans_h_} and \ref{eq:rot_h_}), we use only the hand annotations from the FHB dataset since the four annotated objects in the dataset are not useful for our scenario. To obtain the necessary object annotations, we find the most similar hand pose and configuration in the synthetic dataset and query the pose of a held object.

To obtain the initial estimate of the hand position $p_H$, the hand predictor starts with convolutional layers that operate on a full image $X$. The output is flattened and fed to a set of fully connected layers to regress to $\tilde{p}_H$. 
To estimate the rotation matrix $W_H$ and joint configuration $\beta$, the hand predictor starts with convolutional layers that operate on a hand crop $X_{H}^{c}$. The output is flattened and fed to a set of fully connected layers to regress to $\tilde{W}_H$ and $\tilde{\beta}$.

To estimate the object position $p_O$, the object predictor starts with convolutional layers that operate on full-size images $X_{O}$ of object segments. The output is flattened and concatenated with $p_H$. This vector is passed to a set of fully connected layers and the network outputs $\tilde{p}_O$.
To estimate the object rotation matrix $W_O$ and a shape vector $\delta(X_{O}^{c})$, the object predictor starts with convolutional layers that operate on object segment crops $X_{O}^{c}$. The output is flattened to generate $\tilde{\delta(X}_{O}^{c})$ and an orientation feature vector that is concatenated with $\tilde{W}_H$. To account for potential noisy hand estimates at inference, we add noise to each Euler angle of the hand orientation $W_H$ (max$\pm 30^\circ$). The concatenated vector is fed to a set of fully connected layers to regress to $\tilde{W}_O$.

When training the hand predictor with an object, $p_O$ and $W_O$ are used to refine the initial hand predictions which is achieved by concatenating them with the intermediate hand predictor layers, running them through a set of fully connected layers and outputting the final $\tilde{p_H}'$ and $\tilde{W_H}'$.

\subsubsection{Loss Functions}
To learn the hand pose and configuration we train the network with the {\em mean squared error\/} (MSE) loss between the ground truth and estimated position, rotation matrices and joint angles.

To learn the object position we train the network with the MSE loss between the ground truth and estimated position.
Computing the orientation loss over the whole rotation matrix is challenging due to ambiguities in the representation and object symmetry. However, with a representation that is invariant to symmetry we can facilitate the learning process.
\begin{figure}[t!]
\centering
\scalebox{0.52}{
\begin{tabular}{c|c|c}
$c$ & \multicolumn{1}{c|}{\begin{tabular}[c]{@{}c@{}}Symmetry\end{tabular}} & \multicolumn{1}{c}{\begin{tabular}[c]{@{}c@{}} $R_O$\end{tabular}} \\ \hline
\includegraphics[align=c, width=0.1\textwidth]{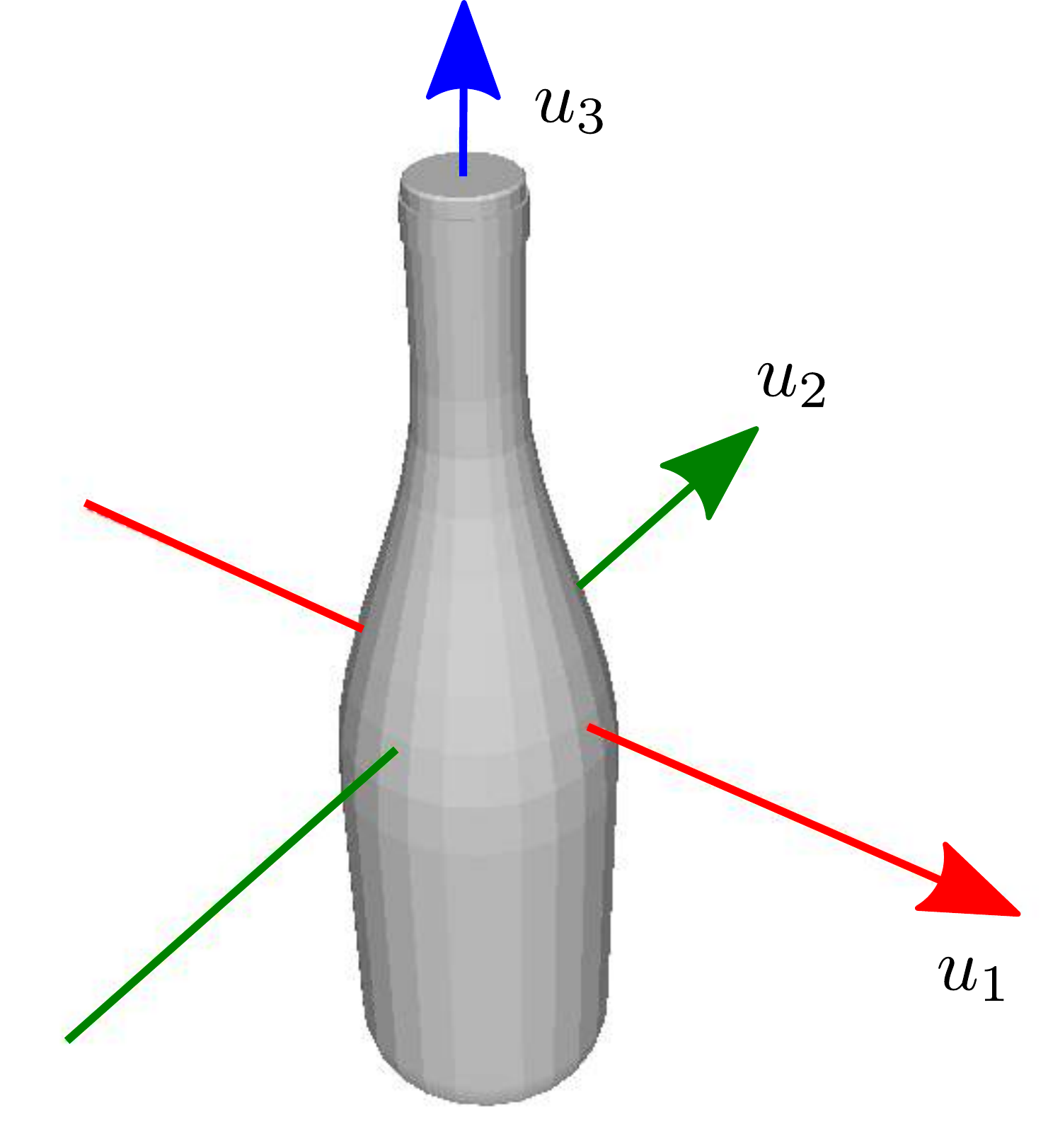} & axial spherical & $R_O = u_3$ \\
\includegraphics[align=c, width=0.1\textwidth]{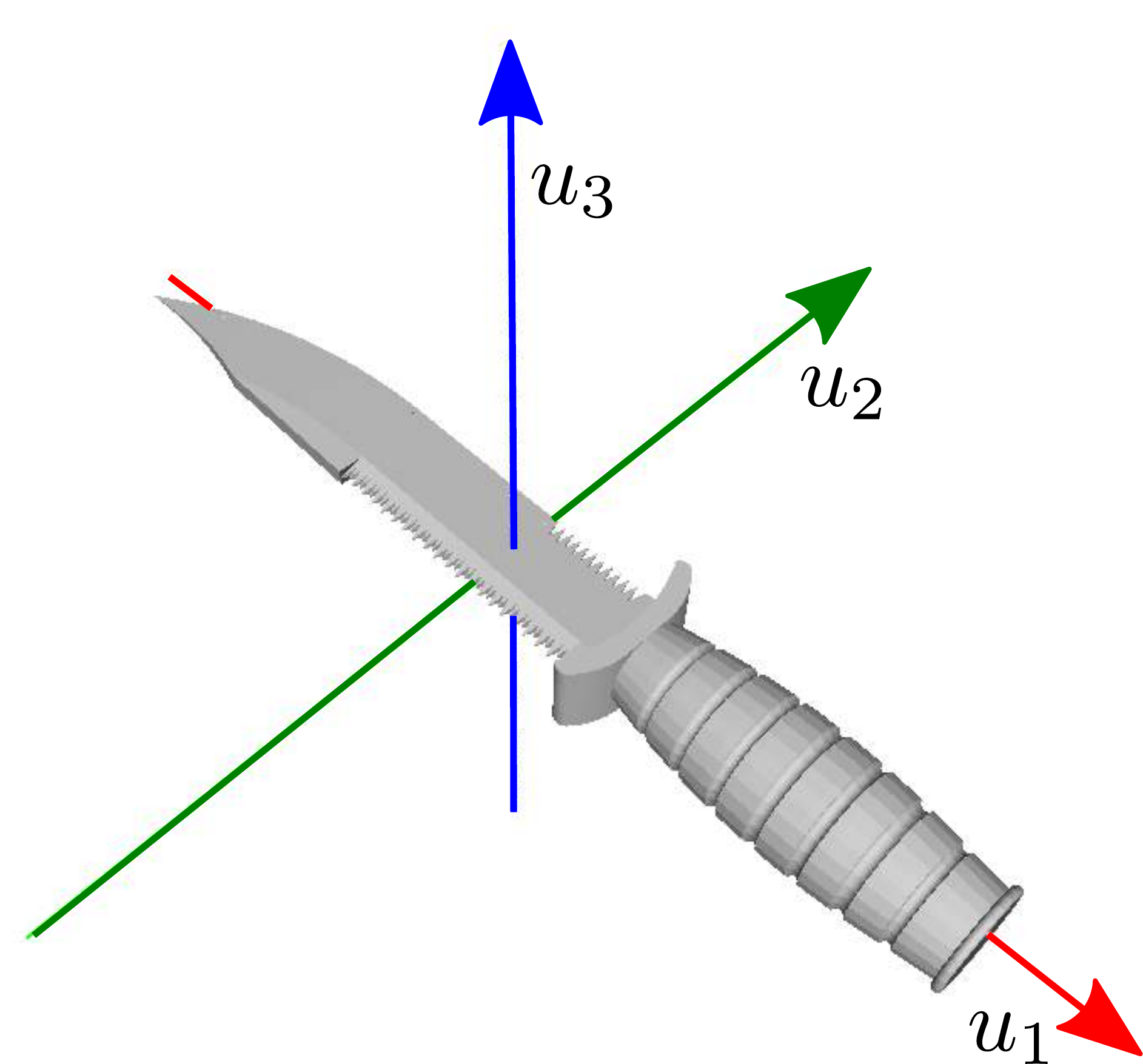} & single plane reflection & $R_O = ([u_1; u_2], u_3 u_3^T)$ \\ 
\includegraphics[align=c, width=0.15\textwidth]{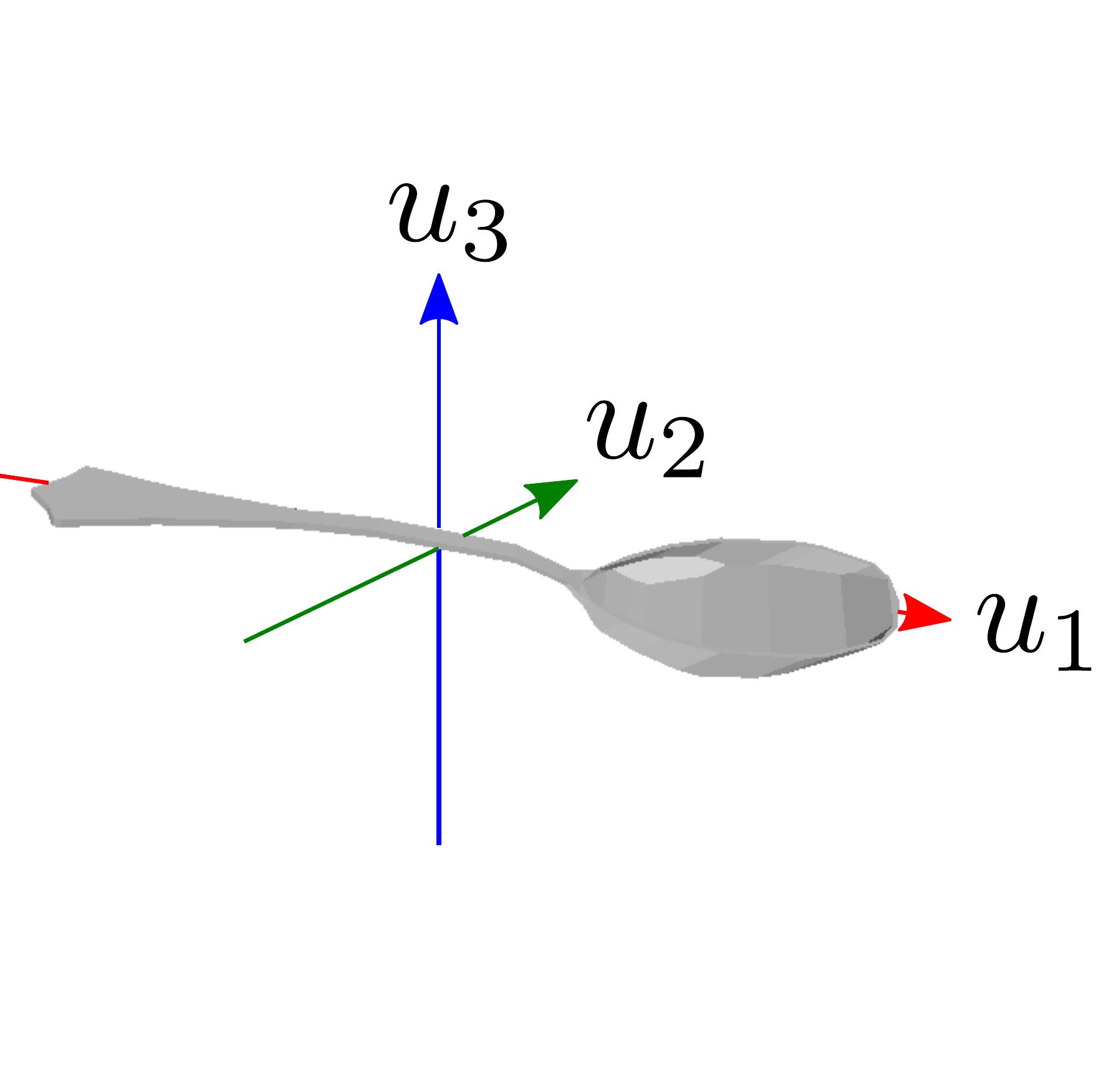} & single plane reflection & $R_O = ([u_1; u_3], u_2 u_2^T)$ \\ 
\end{tabular}
}
\caption{Per category orientation representation $R_O$. It consists of columns $u$ of a rotation matrix $W_O$ and is constructed based on the type of symmetry that the object category possesses.\vspace{-.1cm}}
\label{fig:symmetry}
\end{figure}
Following~\cite{saxena2009learning}, we construct a category-specific representation $R_O$ which consists of columns $u$ of the object rotation matrix and is defined depending on the type of symmetry that an object possesses (see \figref{fig:symmetry}).
For each category, we compute the rotation loss by extracting the columns $\tilde{u}$ from the estimated rotation $\tilde{W}_O$ and constructing $\tilde{R}_O$. The orientation loss is then the MSE between $\tilde{R}_O$ and $R_O$. Therefore, estimated rotations $\tilde{W}_O$ that differ from the ground truth rotations $W_O$ are not penalized if they yield the same $R_O$.

For object shape estimation, we take a retrieval approach that requires to learn an embedding space in which similar data points are mapped close together and distant otherwise. In our model, the first data point is the image crop $X_{O}^{c}$ of the object and the second data point is a mesh $M$ which we represent as a binary occupancy grid denoted by $M_V \in \mathbb{Z}_{2}^{H \times W \times D}$. $X_{O}^{c}$ and $M$ are similar if they belong to the same object.
To learn the embedding space, we construct a Siamese network that outputs the features $\delta(X_{O}^{c})$ and $\delta(M_V)$ and train it with sigmoid cross entropy loss over pairs of features. 

\subsubsection{Inference}
To collect the data for TOG, we run the network on the GUN-71 dataset in which we label each image with a task. 
The full RGB image $X$ is first pre-processed with \emph{Mask R-CNN} to get object segments $X_{O}$, $X_{O}^{c}$ and the category $c$. Simultaneously, we run a hand detector on $X$ to get the hand crop $X_{H}^{c}$. The hand detector is trained on EgoHands~\cite{urooj2018analysis}, a dataset for hands in complex egocentric interactions. This gives us the necessary inputs to the network $\{X, X_{H}^{c}, X_{O}, X_{O}^{c}\}$. 
For each image in the dataset, we first obtain the hand estimates $\{\tilde{\theta}_{H}, \tilde{\beta}\}$, then we use them to get object estimates $\{\tilde{\theta}_{O}, \tilde{\delta({X}_{O}^{c}})\}$. These are finally then used to refine the initial hand estimates. We predict the hand without an object first, since the hand predictor trained in such manner already performs well (in contrast to the object predictor without the hand).
For mesh retrieval at inference time, we rely on pre-computed $\delta(M_V)$ for all training set meshes $\tilde{M}$. To retrieve a suitable object mesh $\tilde{M}$ given $\tilde{\delta({X}_{O}^{c})}$, we compute the probabilities of all $\delta(M_V)$ and $\tilde{\delta({X}_{O}^{c}})$ pairs and retrieve a mesh with highest probability. Finally, we transform the global hand and object poses to the object coordinate frame such that the grasp is encoded with respect to the object. This gives us a set of meshes annotated with human grasps for a certain task.

\vspace{-3px}
\subsection{Task-Oriented Grasping}
\label{sec:inf-tog}
\subsubsection{Network Architecture}
The {\em TOG-T} CNN takes as input a volumetric representation $M_V$ of an object and a grasp $g$. It starts with two 3D convolutional layers to process the mesh and generate a lower dimensional representation of the object shape. This is then concatenated with the grasp vector and passed to a set of fully connected layers that output either $1$ if the grasp is suitable for the task and $0$ otherwise. Furthermore, to account for partial and noisy data in the real world, we add a dropout in the first layer that randomly removes $50\%$ of the input points. The network is trained with sigmoid cross entropy loss. 

\subsubsection{Inference}
Our goal is to execute task-oriented stable grasps with a parallel gripper on a novel object from a known category. 
Concretely, when presented with an object, the robot must reason about what gripper poses with respect to the object ensure stability once the object is lifted and also task suitability in terms of contact locations. This is a difficult problem which poses several challenges. 
First, we need to segment the object from the table and the background. Second, we need to obtain the points on the object surface and their corresponding normals. From this, we can generate the object representation $M_V$ and the grasp representations $g$ that form the input to the {\em TOG-T} network. Finally, we need to plan collision-free grasps and execute them in the robot workspace. 

To segment the object point cloud, we place it in its canonical orientation, next to an April tag whose coordinates in the camera frame are known. 
The segmentation is achieved by filtering the points that are in the proximity to the tag~\cite{varley2017shape}. This gives us a raw object point cloud. 
From the point cloud, we generate the object volumetric representation $M_V$ by scaling the points to fit inside a $50 \times 50 \times 50$ grid. We also compute the normals $n$ which we use together with randomly sampled rolls $\omega$ and offsets $s$ to generate grasps $g$ to be tested. Finally, we run $M_V$ and $g$ through the {\em TOG-T} network to obtain the task-suitable regions and {\em TOG-S} network to obtain the stability scores. 
Finally, we execute the grasp with the highest score that is also suitable for the task and kinematically reachable.
\vspace{-5px}
\section{Experiments}
\label{sec:exp}
In this section, we present experiments and results that test the following hypotheses: 
\begin{enumerate}[wide, labelwidth=!, labelindent=0pt]
{
\item Using a hand helps to facilitate object pose learning and vice versa. In~\secref{sec:eval_quant_a}, we test this by quantitatively comparing object pose estimation results with and without a hand on the synthetic dataset as well as hand pose estimation results with and without an object on the FHB dataset.
\item Representing the hand with a pose and joint configuration instead of 3D joint positions is favorable when requiring precise estimation of object poses. 
In~\secref{sec:eval_quant_b}, we test this hypothesis by comparing object pose estimation results when using either of the two hand representations.
\item Using a combination of real and synthetic data to train our CNN for hand-object estimation yields more accurate results in terms of relative poses than reconstructing the hand-object pairs. We test this  through a qualitative and quantitative comparison of ours and a state-of-the art method for hand-object reconstruction~\cite{hasson2019learning} on the challenging real-world dataset  GUN-71~\cite{rogez2015understanding} (\secref{sec:eval_gun}).
}
\end{enumerate}
In~\secref{sec:eval_tog} we demonstrate that our method can be used to enable a robot to grasp novel objects from a known category in a task-oriented manner. 

\vspace{-4px}
\subsection{Hand-Object Estimation on Synthetic and FHB Datasets}
\label{sec:eval_quant}
\subsubsection{Metrics}
\label{sec:eval_metrics}
In our quantitative experiments we evaluate object pose and shape as well as hand pose and joint configuration.
To evaluate the object shape, we compute the F1 score~\cite{sasaki2007truth} between the ground truth and retrieved shapes in the voxel grid space.
To evaluate the hand configuration we report {\em Mean Absolute Error\/} (MAE) between the ground truth and estimated joint angles in degrees.
To evaluate the poses, we report MAE between the ground truth and estimated orientation (deg) and MAE in position (mm). 

When reporting the orientation errors for objects, we generate all the possible rotation matrices which have the same $R_O$ (see~\secref{sec:hops_obj}), convert them to Euler angles, and compute MAEs between all the generated and the ground truth angles. We report the minimum over all MAEs. Namely, due to symmetry around the $z$-axis in {\em bottles}, all possible rotations around it would yield the same appearance and $R_O$. Therefore, when we report the final orientation MAE, we do not penalize rotations around the $z$-axis that differ from the ground truth. We do similar for {\em knives} and {\em spoons}.

\subsubsection{The Effect of Jointly Estimating Objects and Hands}
\label{sec:eval_quant_a}
\figref{fig:se} shows F1 scores for three categories of objects. We obtain the highest score for {\em bottles}. Furthermore, we report object pose errors \figref{fig:abl_obj} when training with and without a hand. The results show that when using the hand we obtain better results than when omitting the hand. This is because the hand pose informs the object pose estimation.
Overall, we achieve the smallest errors in orientation for {\em bottles}. This is expected since {\em bottles} posses axial spherical symmetry which means we are estimating only one column of its rotation matrix. 
{\em Spoons} achieve the highest error. This is because it is often difficult to discern if the concave or the convex part is facing the camera (which is a strong indication of the orientation).
The same applies vice versa, i.e., training the hand pose estimation with an object yields better results, as shown in~\figref{fig:hand_abl}. We also report mean root-relative end-point error (HP error) (mm) over joint positions and compare our results with~\cite{hasson2019learning}. 

{
\scalebox{.77}{
\begin{minipage}{0.35\textwidth}
    \centering
    \begin{tabular}{ccccc}
    \multicolumn{1}{l}{} & \multicolumn{2}{c}{Position [mm]} & \multicolumn{2}{c}{Orientation [$^\circ$]} \\ \hline
    $c$ & 
    \multicolumn{1}{c}{\begin{tabular}[c]{@{}c@{}}$H$\end{tabular}} & 
    \multicolumn{1}{c}{\begin{tabular}[c]{@{}c@{}}w/o $H$\end{tabular}} &
    \multicolumn{1}{c}{\begin{tabular}[c]{@{}c@{}}$H$\end{tabular}} & 
    \multicolumn{1}{c}{\begin{tabular}[c]{@{}c@{}}w/o $H$\end{tabular}} \\ \hline
    bottle & 19.49 & 58.89 & 11.56 & 17.10\\ 
    knife & 16.00 & 57.98 & 43.92 & 53.79 \\ 
    spoon & 18.28 & 62.25 & 51.23 & 56.44 \\  \hline
    \textbf{avg} & 17.92 & 59.70 & 35.57 & 42.44
    \end{tabular}
    \captionsetup{justification=justified}
    \captionof{figure}{Object pose errors for training of object predictor with and without the hand.}
    \label{fig:abl_obj}
    \end{minipage}
}
\hspace{0.1cm}
\scalebox{0.77}{
\begin{minipage}{0.2\textwidth}
    \centering
    \begin{tabular}{cc}
    \multicolumn{1}{l}{$c$} & \multicolumn{1}{c}{F1 score} \\ \hline
    bottle & 0.88 \\ 
    knife & 0.65  \\ 
    spoon & 0.51 \\  \hline
    \textbf{avg} & 0.68
    \end{tabular}
    \captionof{figure}{F1 scores for shape retrieval. The scores are calculated between $\tilde{M_V}$ and $M_V$.}
    \label{fig:se}
\end{minipage}
}
}

\begin{figure}[ht]
\centering
\scalebox{0.7}{
\begin{tabular}{ccccc}
& Position [mm] & Orientation [$^\circ$] & Joint angles [$^\circ$] & HP error [mm] \\ \hline
$O$ & 131.42 & 19.80 & 13.03 & 45.19\\
w/o $O$ & 146.16 & 20.98 & 13.34 & 46.43\\
\end{tabular}
}
\caption{Hand errors for training the hand predictor with and without an object. We use the same split for training and testing as~\cite{garcia2018first}.\vspace{-.5cm}}
\label{fig:hand_abl}
\end{figure}

\begin{figure}[ht!]
\centering
\scalebox{0.7}{
\begin{tabular}{ccccc}
\multicolumn{1}{l}{} & \multicolumn{2}{c}{Position [mm]} & \multicolumn{2}{c}{Orientation [$^\circ$]} \\ \hline
$c$ & 
\multicolumn{1}{c}{\begin{tabular}[c]{@{}c@{}}$H$-PC\end{tabular}} & 
\multicolumn{1}{c}{\begin{tabular}[c]{@{}c@{}}$H$-JP\end{tabular}} &
\multicolumn{1}{c}{\begin{tabular}[c]{@{}c@{}}$H$-PC\end{tabular}} & 
\multicolumn{1}{c}{\begin{tabular}[c]{@{}c@{}}$H$-JP\end{tabular}} \\ \hline
bottle & 19.49 & 29.36 & 11.56 & 54.82 \\ 
knife & 16.00 & 32.26 & 43.92 & 86.64 \\ 
spoon & 18.28 & 29.76 & 51.23 & 103.84\\  \hline
\textbf{avg} & 18.17 & 30.19 & 35.4 & 81.76
\end{tabular}
}
\caption{Object errors for training the object predictor with two different hand representations: 1) hand pose and joint configuration (H-PC) and 2) 3D joint positions (H-JP).
}
\label{fig:abl_hand_rep}
\end{figure}

\begin{figure*}[ht!]
\centering
\scalebox{0.7}{
\includegraphics[width=\textwidth]{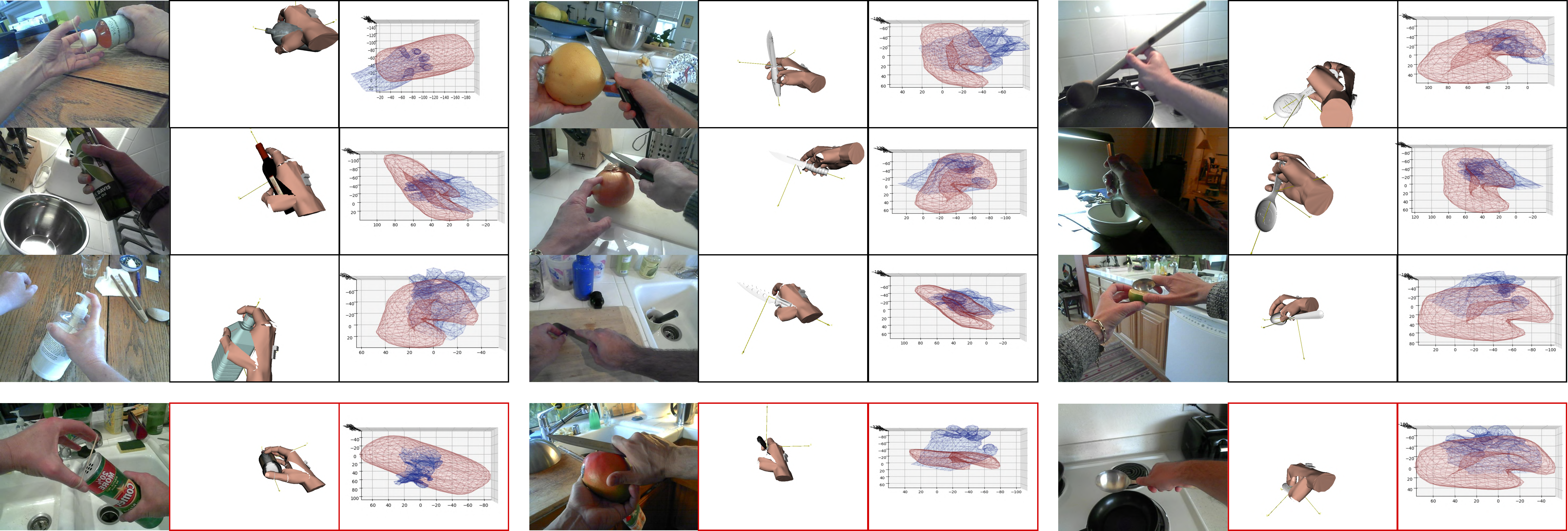}
}
\caption{Qualitative results on GUN-71. We show three examples for \emph{bottle}, \emph{knife} and \emph{spoon}. The second and the third column show results of our method and of the reconstructions from~\cite{hasson2019learning} respectively. Images with the red frame in the bottom row show failures cases on one example per category.}
\label{fig:qual}
\end{figure*}

Our HP error is comparable to state-of-the-art~\cite{hasson2019learning} that achieved the error of $28.82$. In their case, the error is calculated over a filtered FHB dataset (frames where the hand is further than $1$ cm away from the object are omitted). Note that our method operates on hand pose and joint angles and to obtain estimates of joint positions we use inverse kinematics which inevitably induces error. Although estimating hand joint positions instead of pose and configuration would probably reduce this error, we will show in~\secref{sec:eval_quant_b} that hand pose and configuration representation yields better results when the goal is to facilitate object pose estimation.

\subsubsection{The Effect of Different Hand Representations}
\label{sec:eval_quant_b}
We evaluate the effect of two different hand representations on object pose estimation. We report position and orientation MAEs for both cases in~\figref{fig:abl_hand_rep}. Results show that when using the hand pose and joint configuration (H-PC) representation we obtain better results than when using the hand joint positions (H-JP) representation. This is expected when estimating the orientation, since there is a regularity between the hand and the object rotation matrices. This is evident in {\em bottles} which are often grasped such that the $z$-axis of the human hand is perpendicular to the principal axis of the {\em bottle}. On the other hand, high errors in the orientation for H-JP indicate that the network cannot use these values to facilitate object orientation estimation.\vspace{-.2cm}

\begin{figure}
\centering
\scalebox{.7}{
    \begin{tabular}{cccccc}
    \multicolumn{1}{l}{} & \multicolumn{1}{c}{Position [mm]} & \multicolumn{2}{c}{Orientation [$^\circ$]} & \multicolumn{2}{c}{Shape [F1]}\\ \hline
    $c$ & 
    \multicolumn{1}{c}{\begin{tabular}[c]{@{}c@{}}Ours\end{tabular}} & 
    \multicolumn{1}{c}{\begin{tabular}[c]{@{}c@{}}Ours\end{tabular}} & 
    \multicolumn{1}{c}{\begin{tabular}[c]{@{}c@{}}\cite{hasson2019learning}\end{tabular}} &
    \multicolumn{1}{c}{\begin{tabular}[c]{@{}c@{}}Ours\end{tabular}} & 
    \multicolumn{1}{c}{\begin{tabular}[c]{@{}c@{}}\cite{hasson2019learning}\end{tabular}} \\ \hline
    bottle & 36.02 & 23.96 & 63.66 & 0.75 & 0.28\\ 
    knife & 23.59 & 54.35 & 86.32 & 0.80 & 0.08\\ 
    spoon & 27.40 & 36.98 & 86.44 & 0.56 & 0.02\\  \hline
    \textbf{avg} & 29.00 & 38.43 & 78.80 & 0.72 & 0.12
    \end{tabular}
}
\caption{Errors computed on a subset of manually annotated images from GUN-71. For comparison, we report the orientation and shape errors obtained with~\cite{hasson2019learning}.\vspace{-.3cm}}
\label{fig:obj_comp}
\end{figure}
\begin{figure}
\centering
\scalebox{.7}{
    \begin{tabular}{cccccc}
    \multicolumn{1}{l}{} & \multicolumn{1}{c}{Position [mm]} & \multicolumn{2}{c}{Orientation [$^\circ$]} & \multicolumn{2}{c}{Joint angles [$^\circ$]}\\ \hline
    $c$ & 
    \multicolumn{1}{c}{\begin{tabular}[c]{@{}c@{}}Ours\end{tabular}} & 
    \multicolumn{1}{c}{\begin{tabular}[c]{@{}c@{}}Ours\end{tabular}} & 
    \multicolumn{1}{c}{\begin{tabular}[c]{@{}c@{}}\cite{hasson2019learning}\end{tabular}} &
    \multicolumn{1}{c}{\begin{tabular}[c]{@{}c@{}}Ours\end{tabular}} & 
    \multicolumn{1}{c}{\begin{tabular}[c]{@{}c@{}}\cite{hasson2019learning}\end{tabular}} \\ \hline
    hand &  41.49 & 17.34 & 73.40 & 20.96 & 29.43\\ 
    \end{tabular}
}
\caption{Hand errors computed on a subset of manually annotated images from GUN-71 and comparison with~\cite{hasson2019learning}.}
\label{fig:hand_comp}
\end{figure}

\begin{figure*}
\centering
\begin{minipage}{.25\textwidth}
  \centering
  \includegraphics[width=\linewidth]{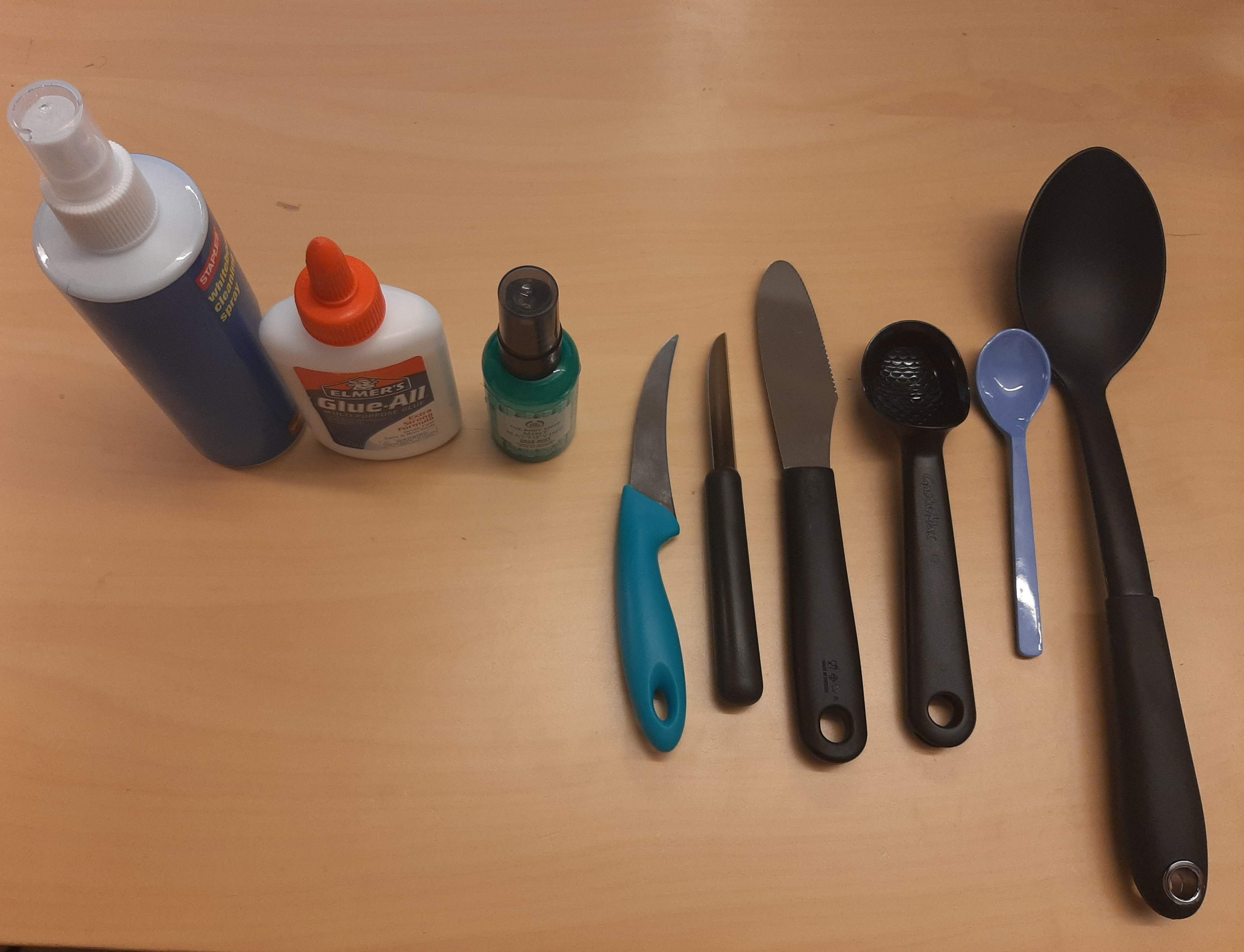}
  \captionof{figure}{We perform task-oriented grasping experiments with a robot on nine nine previously unseen real-world objects; three per category.}
  \label{fig:rwob}
\end{minipage}
\hspace{.5cm}
\begin{minipage}{.55\textwidth}
  \centering  
  \includegraphics[width=\linewidth]{img/robot_exp.pdf}
  \captionof{figure}{Top to bottom: suitable grasping area (green) for pouring with a bottle, cutting with a knife and stirring with a spoon; stability heatmaps (the greener the more stable a grasp); stable, task-oriented grasps on the objects; suitable grasping areas obtained with~\cite{hasson2019learning}.\vspace{-.1cm}}
  \label{fig:robot_exp}
\end{minipage}
\end{figure*}

\subsection{Hand-Object Estimation on a Real-World Dataset}
\label{sec:eval_gun}
\subsubsection{Qualitative Evaluation}
We show a qualitative comparison between our method and~\cite{hasson2019learning} on the GUN-71 dataset in~\figref{fig:qual}. For each category, we show three examples consisting of an input image, results of hand and object estimation using our method and the results of reconstructions. Note that our method operates on full images and outputs global poses in the camera frame while~\cite{hasson2019learning} is trained on a crop around the hand-object pair and outputs reconstructions that are relative to the hand wrist. 

Our method yields more precise object poses and shapes. In comparison,~\cite{hasson2019learning} sometimes reconstructs the wrong object, i.e., instead of the tool it reconstructs the object that the tool is used on. For example, for a knife in the first row of~\figref{fig:qual}, it reconstructs an apple.
This drawback is evident for {\em spoons} and less so for {\em bottles}. 
Furthermore, reconstructed objects lack sufficient detail for learning TOG, e.g., for {\em knives} it is hard to distinguish the blade from the handle or to determine where the sharp side of the blade is. 
This is because: 1)~\cite{hasson2019learning} need to regularize the hand shape which can sometimes fail (i.e., when the hand is occluded by the object) and 2) due to the high dimensionality of the problem, reconstructing an object is more difficult than retrieving its shape and estimating its pose. 

The last row shows failure cases on one example per category which illustrate the limitations of our method. For example, in {\em knives}, it is sometimes difficult to discern the angle of a blade w.r.t. the camera. 
Similarly for {\em spoons}, it is difficult to discern if the concave or the convex side is facing the camera. Wrong estimates of the orientation yield high errors, although the visual appearance is quite similar. However, \cite{hasson2019learning} suffers from similar limitations as the exact orientation is almost impossible to detect from their results.

\subsubsection{Quantitative Evaluation}
\label{sec:eval_rw_quant}
We manually annotate $10\%$ ($\simeq 20$ images per category) of the real-world images and report errors for objects in~\figref{fig:obj_comp} and hands in~\figref{fig:hand_comp}.
We also compare the results to~\cite{hasson2019learning}. Their method outputs hand-object meshes, i.e., a set of 3D vertices whose positions are relative to the hand wrist (global positions are omitted). To evaluate the performance of their method, we need to convert their representation to comply with our metrics. In other words, we need to compute object orientation and shape as well as hand orientation and joint angles from reconstructions. 

To get the object estimates, we compare the reconstructed mesh with the meshes in our synthetic dataset that we randomly rotate. We randomly rotate meshes to generate a database of vertex positions with associated orientation and shape annotations. 
Then, we compute distances between vertex positions of the reconstructed and rotated meshes and select the orientation and shape of the mesh with the lowest distance. Finally, we compute errors between the ground truth and estimated values. 
Similarly for the hands, we compute the orientations and joint angles of the reconstructed hand mesh (joint positions) via inverse kinematics and report errors w.r.t. ground truth.

\vspace{-3px}
\subsection{Task-Oriented Grasping With a Robot}
\label{sec:eval_tog}
Given a novel object and a task, we want to execute stable, task-oriented grasps with a robot. We assume that the object is placed on a table in an upright orientation and that it belongs to one of the three categories: {\em knives, bottles} or {\em spoons}. 
To demonstrate the applicability of our method in a real-world setting, we present a robot with a novel object from a known category,~\figref{fig:rwob}, then do TOG with the ABB Yumi robot,~\figref{fig:robot_exp}.
In the first row of~\figref{fig:robot_exp} we show suitable grasping areas (green) for {\em cutting, stirring} and {\em pouring}, stability heatmaps obtained with {\em TOG-S} (second row) and an execution of a reachable, task-oriented grasp with the highest stability score (third row). 

We also qualitatively compare our method with~\cite{hasson2019learning} (last row). We process GUN-71 with~\cite{hasson2019learning} and train a {\em TOG-T} with the processed data (the network has the same architecture as the one trained on the data obtained with our method). The data contains hand-object poses that we computed by adapting their output to our pose and shape representation as explained in Sec.~\ref{sec:eval_rw_quant}.
Finally, we run the {\em TOG-T} on novel objects and show inferred grasping areas. The areas often include parts that should be avoided if we want a task to succeed, e.g., caps on bottles when pouring. 
Furthermore, such grasps are almost never demonstrated in GUN-71 which suggests that the method does not perform very well when precise estimation of poses is necessary.
\section{Conclusion and Future Work}
We presented an approach for TOG by processing a real-world dataset of RGB images showing hands and objects in interaction. We devised a CNN that takes as input an RGB image and outputs a hand pose and configuration as well as an object pose and a shape descriptor. 
We then used this data to train a CNN that predicts task-suitable grasping regions on novel objects. 
Our ablation study showed that adding a hand to the training of the object predictor improves the object estimates and vice versa. 
Results on GUN-71 demonstrated the competitiveness of our method w.r.t. state-of-the-art and hardware experiments showed that we can use our method to teach a robot to execute task-oriented grasps on novel objects.
In the future, we plan on: i) adding more object categories and ii) estimating hands and objects from video sequences. 
\bibliography{main.bib}

\end{document}